\pdfoutput=1
\documentclass[11pt]{article}

\usepackage{ACL2023}

\usepackage{times}
\usepackage{latexsym}
\usepackage{amssymb}
\usepackage{booktabs}
\usepackage{amsfonts}
\usepackage{caption}
\usepackage{threeparttable}
\usepackage{geometry}
\usepackage{multirow}
\usepackage{microtype}
\usepackage{booktabs}
\usepackage{algorithm}
\usepackage[noend]{algpseudocode}
\usepackage{subcaption}
\usepackage{amsmath}
\usepackage{graphicx}
\usepackage{threeparttable, tablefootnote}
\usepackage[T1]{fontenc}
\usepackage[utf8]{inputenc}

\usepackage{microtype}

\usepackage{inconsolata}
\usepackage{colortbl}
\newcommand*{\affmark}[1][*]{\textsuperscript{#1}}
\makeatletter
\def\thanks#1{\protected@xdef\@thanks{\@thanks
        \protect\footnotetext{#1}}}
\makeatother

\title{Class-Adaptive Self-Training for Relation Extraction \\ with Incompletely Annotated Training Data}

\author{Qingyu Tan\affmark[1, 2]
\thanks{Qingyu Tan and Lu Xu are under the Joint PhD Program between Alibaba and NUS/SUTD. }~~~\textbf{Lu Xu\affmark[1, 3]~~~Lidong Bing\affmark[$^\dag$ 1] \thanks{$^\dag$  Corresponding author.}~~~Hwee Tou Ng\affmark[2]} 
\\$^1$DAMO Academy, Alibaba Group~~\\
$^2$Department of Computer Science, National University of Singapore\\
$^3$Singapore University of Technology and Design \\
\texttt{\{qingyu.tan,lu.x,l.bing\}@alibaba-inc.com}\\
\texttt{\{qtan6,nght\}@comp.nus.edu.sg}\\
}

\begin{document}
\maketitle
\begin{abstract}
Relation extraction (RE) aims to extract relations from sentences and documents. Existing relation extraction models typically rely on supervised machine learning. However, recent studies showed that many RE datasets are incompletely annotated. This is known as the false negative problem in which valid relations are falsely annotated as \textit{no\_relation}. Models trained with such data inevitably make similar mistakes during the inference stage.  Self-training has been proven effective in alleviating the false negative problem. However, traditional self-training is vulnerable to confirmation bias and exhibits poor performance in minority classes. To overcome this limitation, we proposed a novel class-adaptive re-sampling self-training framework. Specifically, we re-sampled the pseudo-labels for each class by precision and recall scores. Our re-sampling strategy favored the pseudo-labels of classes with high precision and low recall, which improved the overall recall without significantly compromising precision. We conducted experiments on document-level and biomedical relation extraction datasets, and the results showed that our proposed self-training framework consistently outperforms existing competitive methods on the Re-DocRED and ChemDisgene datasets when the training data are incompletely annotated\footnote{Our code is released at \url{https://github.com/DAMO-NLP-SG/CAST}}.

\end{abstract}

% Reduce the TACRED content.  Well-annotated dev set, dev set of Re-DocRED 
% Well-annotated dev set removal 

\section{Introduction}
Relation extraction (RE) \cite{wang-etal-2019-tackling, chia-etal-2022-dataset} is an important yet highly challenging task in the field of information extraction (IE). 
Compared with other IE tasks, such as named entity recognition (NER) \cite{xu-etal-2021-better}, semantic role labeling (SRL) \cite{li-etal-2021-syntax}, and aspect-based sentiment analysis (ABSA) \cite{li-etal-2018-hast, zhang-etal-2021-aspect-sentiment}, RE typically has a significantly larger label space and requires graphical reasoning \citep{christopoulou-etal-2019-connecting}. 
The complexity of the RE task inevitably increases the difficulty and cost of  producing high-quality benchmark datasets for this task. 

%\bing{Reduce the content for sentence level RE, say to half.}
In recent years, several works that specifically focus on revising the annotation strategy and quality of existing RE datasets were conducted~(\citealp{stoica2021re}; \citealp{alt2020tacred}; \citealp{tan2022revisiting}).
%Compared to sentence-level relation extraction, the DocRE task is more challenging in the following aspects: 1. The context length for DocRE is significantly higher, and it requires complex cross-sentence reasoning. 2. On average, there are more entities in one example for DocRE compare to its sentence-level counterpart, and the complexity of relation extraction scales quadratically with respect to the number of entities. 
% \bing{quality is not well?}
For example, the DocRED~\citep{yao2019docred} dataset is one of the most popular benchmarks for document-level relation extraction. This dataset is produced by the recommend-revise scheme with machine recommendation and human annotation. However, \citet{Huang2022DoesRP} and \citet{tan2022revisiting} pointed out the false negative problem in the DocRED dataset, indicating that over 60\% of the relation triples are not annotated.
To provide a more reliable evaluation dataset for document-level relation extraction tasks, \citet{Huang2022DoesRP} re-annotated 96 documents that are selected from the original development set of DocRED. 
In addition, \citet{tan2022revisiting} developed the Re-DocRED dataset to provide a high-quality revised version of the development set of DocRED. The Re-DocRED dataset consists of a development set that contains 1,000 documents and a silver-quality training set that contains 3,053 documents. 
Nevertheless, both works on DocRED revision did not provide gold-quality datasets due to the high cost of annotating the relation triples for long documents.
Learning from incompletely annotated training data is crucial and practical for relation extraction. Hence, in this work, we focused on improving the training process with incompletely annotated training data. 
%\bing{do not mention well-annotated Dev set before experiment. Only in experiment, you say you have two datasets, DocRED experiment, we use Re-DocRED dev for a more reliable evaluation conclusion, but you also have the result with a less well-annotated dev. For biomedical dataset, we have to use its less solid dev.}

\begin{figure*}
    \centering
    \resizebox{\textwidth}{!}{
    \includegraphics{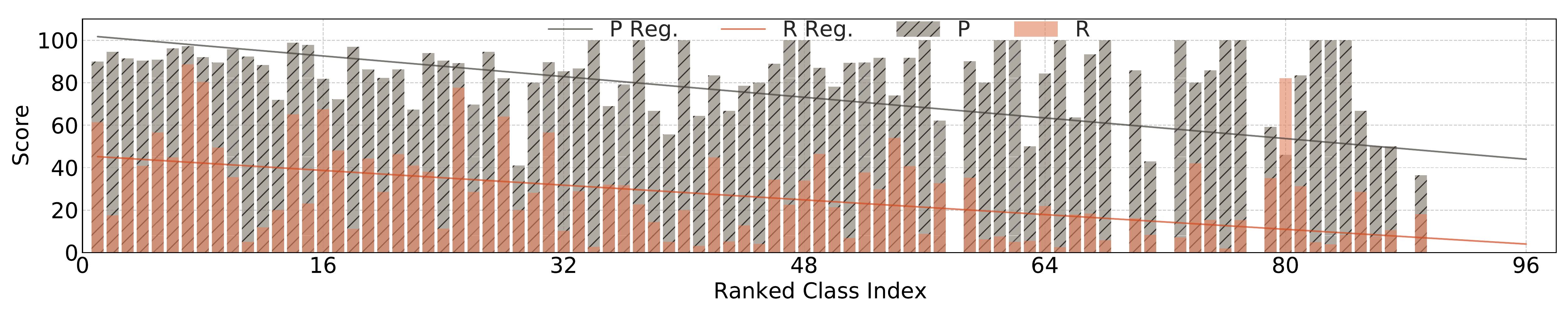}}
     \caption{Precision and recall scores of each class (ranked by class frequency: left [high] \textrightarrow right [low]) on the development set of Re-DocRED when the model is trained on DocRED. P Reg. and R Reg. stand for the regression lines of the scores.}
     %\bing{improve it as discussed.} \textcolor{red}{seems only have 90 classes? or the rest are zeros?}}
    \label{fig:p-r-per-class}
\end{figure*}

%\bing{any citation for the confirmation bias?}
To tackle the problem of training with incompletely annotated datasets, prior works leveraged the self-training method to alleviate the detrimental effects of false negative examples (\citealp{feng2018reinforcement}; \citealp{hu-etal-2021-semi-supervised}; \citealp{chen2021h}; \citealp{zhou-etal-2023-improving}). 
However, self-training-based methods are highly susceptible to confirmation bias, that is, the erroneously predicted pseudo-labels are likely to deteriorate the model's performance in subsequent rounds of training ~(\citealp{arazo2020pseudo}; \citealp{tarvainen2017mean}; \citealp{Li2020DivideMix}). 
Furthermore, the label distribution of relation extraction task is highly imbalanced. 
Therefore, the predictions made by prior self-training methods are likely to be of the majority classes. \citet{wei2021crest} proposed a re-sampling strategy based on class frequencies to alleviate this problem in image classification. 
In this way, not all generated pseudo-labels will be used to update the training datasets.
%\bing{preserved or reserved?}
The pseudo labels of the minority classes have higher probabilities to be preserved than those of the frequent classes. 
However, such a sampling strategy does not specifically address the problems caused by the erroneously generated pseudo labels.
When a model is trained on incompletely annotated datasets, minority classes exhibit bad performance and  frequent classes may have low recall scores, as shown in Figure \ref{fig:p-r-per-class}. 
Merging pseudo labels with original labels of the training dataset without considering the correctness of the former potentially deteriorates performance in subsequent iterations. 
% This problem is particularly harmful for minority classes because these classes usually have low initial prediction quality and a small number of false positive examples from self-training can brings in a big impact on performance in the next interaction. 

In order to overcome confirmation bias in self-training, we proposed a class-adaptive self-training (\textbf{CAST}) approach that considers the correctness of the pseudo labels. 
Instead of sampling the pseudo labels based on class frequencies, we introduced a class-adaptive sampling strategy to determine how the generated pseudo labels should be preserved. 
Specifically, we calculated the precision and recall scores of each class on the development set and used the calculated scores to compute the sampling probability of each class.
Through such an approach, CAST can alleviate confirmation bias caused by erroneous pseudo labels.
Our proposed approach preserves the pseudo labels from classes that have high precision and low recall scores and penalizes the sampling probability for the pseudo labels that belong to classes with high recall but low precision scores.
% Our approach encourages preserving the pseudo labels from the classes that have high precision and low recall scores.
% At the same time, our approach also penalizes the sampling probability for the pseudo labels that belong to the classes of high recall but low precision. 

Our contributions are summarized as follows. (1) We proposed CAST, an approach that considers the correctness of generated pseudo labels to alleviate confirmation bias in the self-training framework. (2) Our approach was evaluated with training datasets of different quality, and the experimental results demonstrated the effectiveness of our approach. (3) Although our approach is not specifically designed for favoring the minority classes, the minority classes showed more significant performance improvements than the frequent classes, which is a nice property as the problem of long-tail performance is a common bottleneck for real applications.
%\bing{this paragraph needs a big improvement to make our contributions clear and convincing.}

\section{Related Work}
\label{sec:related-work} 
%\bing{talk less about sentence-level RE.}
\paragraph{Neural Relation Extraction} Deep neural models are successful in sentence-level and document-level relation extraction. \citet{zhang-etal-2017-position} proposed position-aware attention to improve sentence-level RE and published TACRED, which became a widely used RE dataset.
\citet{yamada-etal-2020-luke} developed LUKE, which further improved the SOTA performance with entity pre-training and entity-aware attention. \citet{chia-etal-2022-relationprompt} proposed a data generation framework for zero-shot relation extraction. However, most relations in real-world data can only be extracted based on inter-sentence information.
To extract relations across sentence boundaries, recent studies began to explore document-level RE. 
As previously mentioned, \citet{yao2019docred} proposed the popular benchmark dataset DocRED for document-level RE. \citet{zeng-etal-2020-double} leveraged a double-graph network to model the entities and relations within a document. To address the multi-label problem of DocRE, \citet{zhou2021document} proposed using adaptive thresholds to extract all relations of a given entity pair. 
\citet{zhang2021document} developed the DocUNET model to reformulate document-level RE as a semantic segmentation task and used a U-shaped network architecture to improve the performance of DocRE. \citet{tan2022document} proposed using knowledge distillation and focal loss to denoise the distantly supervised data for DocRE and achieved great performance on the DocRED leaderboard. 
However, all preceding methods are based on a closed-world assumption (i.e., the entity pairs without relation annotation are negative instances). This assumption ignores the presence of false negative examples. Hence, even the above-mentioned state-of-the-art methods may not perform well when the training data are incompletely annotated.

%Most recently, \citet{tan2022revisiting}  showed that the annotation of the DocRED dataset is highly incomplete, and it is not a fair reference for the document-level RE task.      
\paragraph{Denoising for Relation Extraction} RE is susceptible to noise in the training data. Noisy data can be categorized into two types: false positives (FPs) and false negatives (FNs). False positive examples are mainly caused by misalignment of knowledge bases. 
\citet{xiao-etal-2020-denoising} proposed a denoising algorithm that filters FP examples in distantly supervised data. \citet{wang-etal-2019-tackling} tackled the class-imbalance problem of RE and NER by meta-learning. The false negative problem is also common in information extraction. \citet{li2020empirical, xu-etal-2023-sampling} used simple negative sampling strategies to alleviate the detrimental effects of FN examples on NER. Most recently, \citet{guo-etal-2023-towards} tackled the multi-label problem in RE by entropy minimization and supervised contrastive learning. Given that the FN problem is related to incomplete annotation, supplementing the annotation by self-training is a viable way to tackle this problem (\citealp{erkan-etal-2007-semi}; \citealp{sun-etal-2011-semi}; \citealp{chen2021h}; \citealp{hu-etal-2021-semi-supervised}). However, self-training is susceptible to confirmation bias; conventional self-training suffers from the problem of error propagation and makes overwhelming predictions for frequent classes. 
Prior research on semi-supervised image classification (\citealp{wei2021crest}; \citealp{he2021rethinking}) indicated that re-sampling of pseudo-labels can be beneficial to class-imbalanced self-training. 
However, existing re-sampling strategies are dependent only on the frequencies of the classes and do not consider the actual performance of each class. 
Our method alleviates confirmation bias by employing a novel re-sampling strategy that considers the precision and recall of each class on the development set. In this way, we can downsample the predictions for popular classes and maintain high-quality predictions for long-tail classes.

%\paragraph{Class-imbalanced } Class-imbalanced self-training has gained increasing attention for its relevance to real-world applications. 

\section{Methodology}
\subsection{Problem Definition}
\label{sec:prob-definition}

Document-level relation extraction (DocRE) is defined as follows: given a text $T$ and a set of $n$ entities $\{e_1, ..., e_n\}$ appearing in the text, the objective of the document-level RE is to identify the relation type $r \in C \cup \{\textit{no\_relation}\}$ for each entity pair $(e_i, e_j)$. 
Note that $e_i$ and $e_j$ denote two different entities, and $C$ is a predefined set of relation classes.
% , including \textit{no\_relation}.
The complexity of this task is quadratic in the number of entities, and the ratio of the NA instances (\textit{no\_relation}) is very high compared with sentence-level RE. Therefore, the resulting annotated datasets are often incomplete.
The setting of this work is to train a document-level RE model with an incompletely labeled training set, and then the model is evaluated on a clean evaluation dataset, such as Re-DocRED~\citep{tan2022revisiting}. 

We denote the training set as $S_T$ and the development set as $S_D$. Two types of training data are used in this work, each representing a different annotation quality. The first type is the training split of the original DocRED data \citep{yao2019docred}, which we refer to as bronze-level training data. 
This data is obtained by a recommend-revise scheme. Even though the annotation of this bronze level is precise, there are a significant number of missing triples in this dataset. 
On the other hand, the training set of the Re-DocRED dataset has added a considerable number of triples to the bronze dataset, though a small number of triples might still be missed. We refer to this Re-DocRED dataset as silver-quality training data. 

%\paragraph{}\noindent
%Since the aim of this work is to explore a better re-sampling strategy when applying self-training to an incompletely annotated dataset, it is necessary to use development and test sets with high quality, which allows us to draw more reliable conclusions. 
%Specifically, the development and test sets of the SentRE and DocRE experiments are from Re-TACRED and Re-DocRED, respectively.
%Details of the dataset statistics can be found in Table~\ref{tab:data-stats}. 
% \bing{should also mention DocRE dev and test sets from Re-DocRED}

\begin{figure}[t]
    \centering
    \resizebox{\columnwidth}{!}{
    \includegraphics[width=\columnwidth]{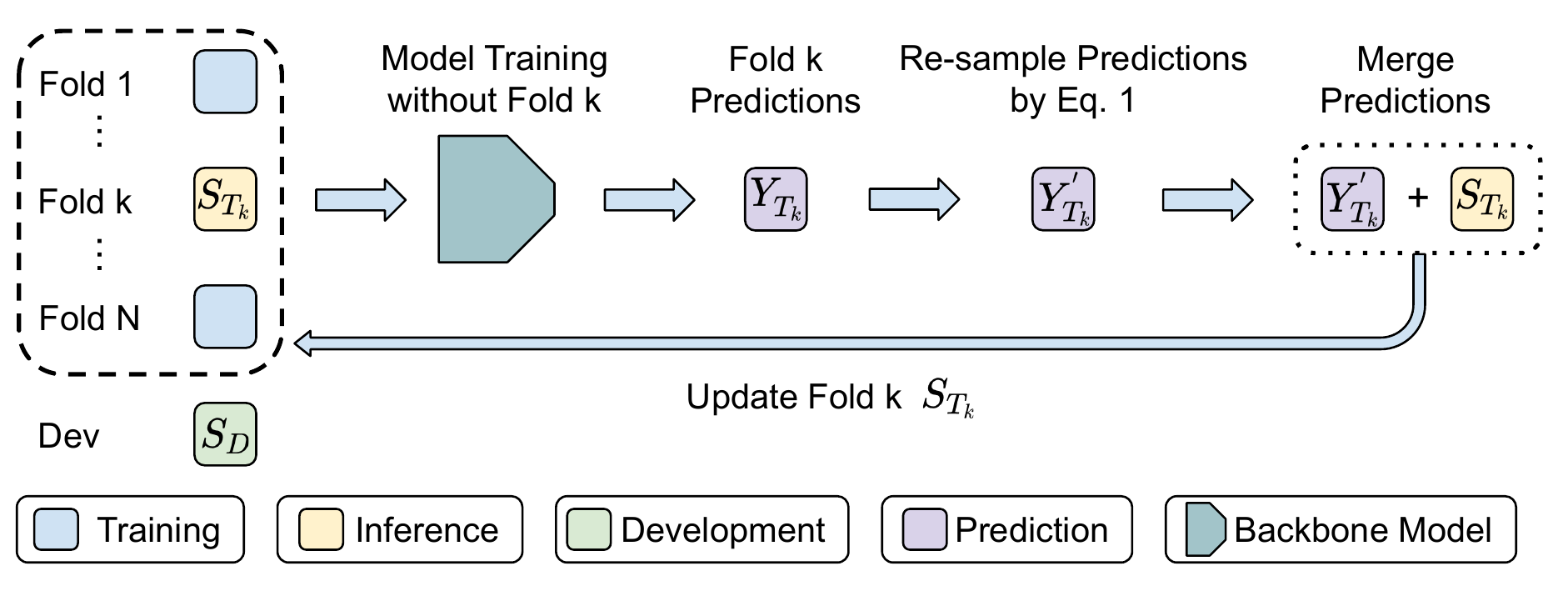}}
    \caption{Illustration of training dataset update of CAST, and Algorithm \ref{alg:cap} describes its full details.}
    \label{fig:model-fig}
\end{figure}
%\bing{put $S_D$ in the figure.}

\subsection{Our Approach}
\label{sec:approach}

\subsubsection{Overview}
The main objective of our approach is to tackle the RE problem when the training data $S_T$ is incompletely annotated. 
We propose a class-adaptive self-training (CAST) framework, as shown in Figure~\ref{fig:model-fig}, to pseudo-label the potential false negative examples within the training set. 
First, we split the training set into $N$ folds and train an RE model with $N-1$ folds.
The remaining  fold $S_{T_k}$ is used for inference. 
Next, we use a small development set $S_D$ to evaluate the models and calculate the sampling probability for each relation class (Eq. \ref{eqn:p-r-prob}). 
The predicted label set $Y_{T_k}$ is obtained by conducting inference on $S_{T_k}$. 
Then, we re-sample the predicted labels based on the computed probability, which is calculated 
% The predicted labels will then be re-sampled based on the above probability, which is calculated 
based on the performance of each class. 
The re-sampled label set is denoted as $Y^{'}_{T_k}$. 
Lastly, $Y^{'}_{T_k}$ will be merged with the initial labels of $S_{T_k}$. 
The details of the proposed framework are discussed in the following subsections.
% We will further elaborate our framework in the following sections.
%\bing{a clear workflow diagram is needed !!}

%We tackle this problem in two aspects, the first one is \textcolor{red}{discarding/relabeling} for the false negative examples, which is achieved by our class-adaptive self-training framework. The second part of this problem is to find harder negative examples to better learn from the confusing examples.

%\bing{need a citation}

\subsubsection{Self-Training}
In traditional self-training, models are trained on a small amount of well-annotated data and pseudo-labels are generated on unlabeled instances \citep{Zhu2009IntroductionTS}.
However, we do not have access to well-annotated training data, and our training data contains false negative examples. 
Therefore, we need to construct an $N$-fold cross-validation self-training system. 
Given a set of training documents $S_T$ with relation triplet annotation, these documents are divided into $N$ folds. 
The first $N-1$ folds will be used for training an RE model. Then, the trained model will be used to generate pseudo-labels for the held out $N$-th fold. 
The pseudo-labels will be merged with the original labels, and the merged data will be used to train a new model. 
The $N$-fold pseudo labeling process will be repeated for multiple rounds until
no performance improvement is observed on the final RE system.
However, because the class distribution of the document-level RE task is highly imbalanced, pseudo-labeling  may favor the popular classes during prediction. This inevitably introduces large confirmation bias to popular classes, which is similar to the ``rich-get-richer'' phenomenon \cite{cho-rich-richer}.

%\bing{not accurate.}
\subsubsection{Intuition} When the annotation of the training set is incomplete, the model trained on such data typically shows high precision and low recall scores for most of the classes. 
Figure~\ref{fig:p-r-per-class} shows the precision and recall of each class of the model that is trained on the DocRED dataset and evaluated on the development set of Re-DocRED. Among the 96 classes, most of the classes obtain higher precision scores than recall scores.
Only one class that has a higher recall score than precision score; some classes have 0 precision and recall scores. 
Given this empirical observation, 
boosting self-training performance by sampling more pseudo-labeled examples from the classes that have high precision and low recall is a good strategy because (1) the pseudo labels of such classes tend to have better quality and (2) the recall performance of these classes can be improved by adding true positive examples. 
For extreme cases in which a class has predictions that are all wrong (i.e. its precision and recall are both 0), the logical action is to discard the corresponding pseudo-labels.

\subsubsection{Class-Adaptive Self-Training (CAST)}
As previously mentioned, traditional self-training suffers from confirmation bias, especially for RE task that has a highly imbalanced class distribution. 
The pseudo-labels that are generated by such an approach tend to be biased toward the majority classes. 
To alleviate this problem, we propose a class-adaptive self-training framework that filters the pseudo-labels by the per-class performance. 
Unlike existing self-training re-sampling techniques (\citealp{wei2021crest}; \citealp{he2021rethinking}) that take only the class frequencies into account, our framework samples  pseudo-labels based on their performance on the development sets. 

First, we evaluate the model for pseudo-labeling on the development set $S_D$ and calculate the precision $P$ and recall $R$ for each class.
%\bing{I think the following two equations are not needed.}
\begin{algorithm}[t]
\caption{Class-Adaptive Self-Training}\label{alg:cap}
\begin{algorithmic}
\scriptsize
\State \textbf{Input}: 
\State $M$: Number of rounds 
\State $N$: Number of folds
\State $S_{T}$: An incompletely annotated training set
\State $S_{D}$: A task-specific development set
\State $\theta$: A backbone model with parameters
%\State $\alpha$: Precision coefficient
\State $\beta$: Smoothness coefficient
\\\hrulefill
\For{$j \in \{1, .., M\}$}
\State $S_{T} = \{S_{T_{1}},...,S_{T_{N}}\}$ \Comment{Split $S_{T}$ into $N$ folds}
\For{$k \in \{1, .., N\}$}
\State $S_{T^{-}} \gets S_{T} - S_{T_{k}}$ 
\State $\theta^{*}_{k} \gets $ Optimize $\theta_{k}$ with $(S_{T^{-}}, S_{D})$
\Comment{Training w/o $S_{T_{k}}$}
\State $Y_{T_{k}} \gets $ Inference on $S_{T_{k}}$ by $\theta^{*}_{k}$
\Comment{Predict labels of $S_{T_{k}}$}
\State Compute $P_i$, $R_i$ with $\theta^{*}_{k}$ and $S_{D}, \forall i \in C$
\State $\mu_{i} \gets [P_{i} * (1 - R_{i})]^{\beta}, \forall i \in C$
\Comment{Eq. 1}
\State $Y^{'}_{T_{k}} \gets $ Re-sample $Y_{T_{k}}$ with rates $\{\mu_i\}, \forall i \in C$ 
\State $S^{'}_{T_{k}} \gets $ Merge $Y^{'}_{T_{k}}$ with annotation of $S_{T_{k}}$

\EndFor
\State $S_{T} \gets S^{'}_{T_{1}}\cup...\cup S^{'}_{T_{N}}$
\Comment{Update training set} 
\State $\theta^{*}_{j} \gets $Optimize $\theta$ with $(S_{T}, S_{D})$
\Comment{Save model for round $j$}
\EndFor
\State $\theta^{*} \gets $ evaluate $\{\theta^{*}_{1},...,\theta^{*}_{M}\}$ on $S_{D}$

\State \textbf{return} $\theta^{*}$
\end{algorithmic}
\end{algorithm}
Then, we define our sampling probability $\mu_{i}$ for each relation class $i$ as:
\begin{equation}
    \mu_{i} = [P_{i} * (1 - R_{i})]^{\beta}
\label{eqn:p-r-prob}
\end{equation}
% \begin{equation}
%     \mu_{i} = (P_{i})^{\beta} * (1 - R_{i})^{\beta}
% \label{eqn:p-r-prob}
% \end{equation}
where $P_{i}$ and $R_{i}$ are the precision and recall scores of class $i$, respectively, and $\beta$ is a hyper-parameter that controls the smoothness of the sampling rates. 
% \textcolor{red}{include the formulation for precision and recall, it might be easier to understand the algo.} 

Note that all pseudo labels will be used when the sampling probability equals to 1. Conversely, all the pseudo labels will be discarded when the sampling probability equals to 0. 
If the recall of a specific class is very small and its precision is close to 1, the sampling rate of the class will be closer to 1. On the contrary, if the recall for a certain class is high, the sampling rate of the class will be low. In this way, our method is able to alleviate confirmation bias toward the popular classes, which typically have higher recall. 
The pseudo-code of our proposed CAST framework is provided in Algorithm~\ref{alg:cap}.

%\subsection{Hard Negative Mining}
%\bing{delete HNM? decide later}
%Our class-adaptive self-training aims to re-label the false negative examples to positive ones. At the same time, we want to emphasize on the more challenging negative examples (entity pairs that have no relation). For a given negative entity pair $(s,o)$, where $s$ stands for subject and $o$ stands for object, we define the uncertainty score for an entity pair with an entropy-like formulation:
%\begin{equation}
%    U = -\sum_{i=1}^{|C|}P(r=i| s,o)log(P(r=i| s,o))
%\end{equation} 
%\bing{directly say it is the entropy over the positive classes.}
%where $U$ stands for the entropy over all positive classes, $|C|$ is the number of relations and $P(r=i| s,o)$ is the probability of entity pair $(s,o)$ having relation $i$. After we rank the negative instances, we select the top $\lambda$ percent of the negative instances during training. These negative instances generally have higher probabilities for all the positive classes, hence they are more confusing compared  
%\bing{also need to say, why such negative examples are hard?} 
%\bing{also need to say how HNM works together with the self-training.}

\begin{table}[t]
\centering
\resizebox{\columnwidth}{!}{
\begin{tabular}{@{}l@{}c@{~~}c@{~}c@{~}c@{}} 
\toprule
\multicolumn{1}{l}{\multirow{2}{*}{\textbf{DocRE}}}  & \textbf{DocRED} & \textbf{Re-DocRED} & \multicolumn{2}{c}{\textbf{\textbf{Re-DocRED}}}  \\
\multicolumn{1}{c}{}                                 & \textbf{Train}  & \textbf{Train}     & \textbf{Dev} & \textbf{Test}                     \\ 
\midrule
\textbf{\# Documents}                                & 3,053           & 3,053              & 500          & 500                               \\
\textbf{Avg. \# Entities per Doc}                            & 19.4            & 19.4               & 19.4         & 19.6                              \\
\textbf{Avg. \# Triples per Doc}                             & 12.5            & 28.1               & 34.6         & 34.9                              \\
\textbf{Avg. \# Sentences per Doc}                           & 7.9             & 7.9                & 8.2          & 7.9                               \\
\textbf{\# NA rate}                                  & 97.0\%          & 94.3\%             & 93.1\%       & 93.1\%                            \\ 
\toprule
\toprule
\multicolumn{1}{l}{\multirow{2}{*}{\textbf{BioRE}}} &  & \multicolumn{3}{c}{\textbf{\textbf{ChemDisGene}}}  \\
\multicolumn{1}{c}{}                                 &  & \textbf{Train}     & \textbf{Dev} & \textbf{Test}                     \\ 
\midrule
\textbf{\# Documents}                                 &           &     76,544       & 1,480       &       523                      \\
\textbf{Avg. \# Words}                            &             & 196.6               & 237.3         & 235.6 \\
\textbf{Avg. \# Entities per Doc}                            &            & 7.6               & 9.0        & 10.0                              \\
\textbf{Avg. \# Triples per Doc}                             &             & 2.2               & 2.2         & 7.2                              \\
\textbf{Avg. \# Sentences per Doc}                           &              & 12.6                & 14.0          & 13.2                               \\
\textbf{\# NA rate}                                  &          & 96.8\%             & 97.7\%       & 93.8\%                            \\ 
\bottomrule
\end{tabular}}
\caption{Dataset statistics of our experiments for DocRE and BioRE. }
\label{tab:data-stats}
\end{table}

\section{Experiments}

\subsection{Experimental Setup}

Our proposed CAST framework can be applied with any backbone RE model. 
For the experiment on DocRED, we adopted the ATLOP \citep{zhou2021document} model as the backbone model, which is a well-established baseline for the DocRE task. We used BERT-Base \citep{devlin-etal-2019-bert} and RoBERTa-Large \citep{liu2019roberta} as the encoders. 
In addition to DocRED, we conduct experiments on ChemDisGene \citep{zhang-etal-2022-distant}, a DocRE dataset for biomedical relation extraction (BioRE). We used the PubMedBERT \citep{gu2021domain} encoder for the BioRE experiments.
We use the development set of Re-DocRED in the document-level RE experiments because the Re-DocRED dataset has a high quality. 
Moreover, we use the distantly-supervised development set of ChemDisGene for the BioRE experiments. Our final models are evaluated on the test sets of Re-DocRED and ChemDisGene. Both of the test sets are human-annotated and have high quality, the statistics of the datasets can be found in Table~\ref{tab:data-stats}. 

For the hyper-parameters, we set $M=5$ (i.e., the iteration round in Algorithm \ref{alg:cap}) and $N=5$ for the self-training-based methods because these methods typically reach the highest performance before the fifth round and five-fold training is the conventional practice for cross validation. For $\beta$, we grid searched $\beta \in \{0.0, 0.25, 0.5, 0.75, 1\}$.
For evaluation, we used micro-averaged F1 score as the evaluation metric. We also evaluate the F1 score for frequent classes and long-tail classes, denoted as Freq\_F1 and LT\_F1, respectively. 
For the DocRED dataset, the frequent classes include the top 10 most popular relation types\footnote{They cover 59.4\% of the positive instances.} in the label space; the rest of the classes are categorized as the long-tail classes. Following \citet{yao2019docred}, we use an additional metric Ign\_F1 on the DocRE task. 
This metric calculates the F1 score for the triples that do not appear in the training data.

%\bing{need to mention some implementation details for ChemDisGene, such as PubMedBERT}

% \textcolor{red}{indicate major hyperparameters ($\beta \lambda$ etc.) here and other not important hyperparameters in the appendix (if no space).}

\begin{table*}[t]
\centering
\resizebox{0.85\textwidth}{!}{
\begin{threeparttable}
\begin{tabular}{llcccccc} 
\toprule
          &\textbf{Model}             & \multicolumn{1}{c}{\textbf{P}} & \multicolumn{1}{c}{\textbf{R}} & \multicolumn{1}{c}{\textbf{F1}} & \multicolumn{1}{c}{\textbf{Ign\_F1}} & \textbf{Freq\_F1}                  & \textbf{LT\_F1}                     \\ 
\midrule
\multirow{7}{*}{\rotatebox[origin=c]{90}{\textbf{BERT}} }
& GAIN$^{\dag}$                   & 88.11                 & 30.98                 & 45.82                  & 45.57                       &  -                         &     -                       \\

& ATLOP        & \textbf{88.39} \small{$\pm$0.39}                 & 28.87 \small{$\pm$0.34}                 & 43.52 \small{$\pm$0.25}                  & 43.28 \small{$\pm$0.24}                    & 45.49 \small{$\pm$0.24} &  40.46 \small{$\pm$0.28}  \\
& SSR-PU-ATLOP$^{\dag}$ & 65.10 \small{$\pm$0.90}                 & 50.53 \small{$\pm$0.89}                 & 56.84 \small{$\pm$0.72}                  & 55.45 \small{$\pm$0.59}                       &         60.21 \small{$\pm$0.64}                     &      51.84 \small{$\pm$0.82}                     \\
& NS-ATLOP     & 74.79 \small{$\pm$0.31}                 & 46.33 \small{$\pm$0.34}                 & 57.22\small{$\pm$0.25}                  & 56.28 \small{$\pm$0.21}                       & 59.23 \small{$\pm$0.23} & 54.13 \small{$\pm$0.24}  \\
& VST-ATLOP    & 63.53 \small{$\pm$1.17}		 & \textbf{56.41} \small{$\pm$0.86}		  & 59.56 \small{$\pm$0.16}	   &58.03 \small{$\pm$0.25}	        &          	63.17 \small{$\pm$0.46}	              &            55.61 \small{$\pm$0.25}                  \\
& CREST-ATLOP  & 69.34 \small{$\pm$1.55}                 & 50.58 \small{$\pm$1.35}                & 58.48 \small{$\pm$0.30}                  & 57.33 \small{$\pm$0.21}                       & 60.31 \small{$\pm$0.64} & 56.33 \small{$\pm$0.15}  \\
\cmidrule(lr){2-8}
& \textbf{CAST-ATLOP} (Ours)   & 70.49 \small{$\pm$1.12}                 & 54.34 \small{$\pm$1.07}                 & \textbf{61.36} \small{$\pm$0.67}                 & \textbf{60.16} \small{$\pm$0.79}                       & \textbf{63.66} \small{$\pm$0.44} & \textbf{58.12} \small{$\pm$0.36}  \\ 
\midrule
\multirow{8}{*}{\rotatebox[origin=c]{90}{\textbf{RoBERTa}} }
& DocuNET$^{\dag}$                & \textbf{94.16}                 & 30.42                 & 45.99                  & 45.88                       & -                           &        -                    \\
& KD-DocRE$^{*}$               & 92.08                 & 32.07                 & 47.57                  & 47.32                       &     -                      &          -                  \\
& ATLOP          & 92.62 \small{$\pm$0.35}	                & 33.61 \small{$\pm$0.48}	                 & 49.32 \small{$\pm$0.29}	              & 49.16 \small{$\pm$0.27}	                       &        51.49 \small{$\pm$0.51}	                   &      45.36 \small{$\pm$0.43}                       \\
& SSR-PU-ATLOP$^{\dag}$   & 65.71 \small{$\pm$ 0.28}                 & 57.01 \small{$\pm$0.47}                 & 61.05 \small{$\pm$0.21}                  & 59.48 \small{$\pm$0.18}                       &     62.85 \small{$\pm$0.10}                      &      58.19 \small{$\pm$0.54}                      \\
& NS-ATLOP       & 68.39 \small{$\pm$2.23}	  & 56.05 \small{$\pm$0.98}	  & 61.58 \small{$\pm$0.48} & 	60.43 \small{$\pm$0.55}       &        	65.35 \small{$\pm$0.12}	                 &          57.16 \small{$\pm$0.44  }                 \\
& VST-ATLOP      & 62.85 \small{$\pm$0.48}                 & \textbf{63.58} \small{$\pm$0.62}                & 63.21 \small{$\pm$0.39}                 & 61.83 \small{$\pm$0.41}                      & 65.68 \small{$\pm$0.43} & 60.09 \small{$\pm$0.45}  \\
& CREST-ATLOP    & 73.09 \small{$\pm$0.79}                 & 55.06 \small{$\pm$0.86}                 & 62.81 \small{$\pm$0.35}                  & 61.90 \small{$\pm$0.33 }                      & 63.71 \small{$\pm$0.41}& 61.75 \small{$\pm$0.49}  \\
\cmidrule(lr){2-8}
& \textbf{CAST-ATLOP }(Ours)    & 72.83 \small{$\pm$0.50}                 & 59.22 \small{$\pm$0.61}                 & \textbf{65.32} \small{$\pm$0.22}                  & \textbf{64.25} \small{$\pm$0.15}                       & \textbf{66.99} \small{$\pm$0.29} &  \textbf{63.05} \small{$\pm$0.11}  \\
\hline
\end{tabular}
  % \begin{tablenotes}[para]
  %   \item[1] \citet{wang2022unified}  \item[2] \citet{tan2022revisiting}
  %   % \vspace{-0.3cm}
  % \end{tablenotes}
\end{threeparttable}
}
\caption{Experimental results on the test set of Re-DocRED when trained with DocRED. Model selection is based on the dev set of Re-DocRED. The reported results are the average of five runs. $^{\dag}$: The results are reproduced from \citet{wang2022unified} with the same development set $S_{D}$.
    $^*$: The results are retrieved from \citet{tan2022revisiting}.}
\label{tab:bronze-redocred}
\end{table*}
%\bing{why some results are missing? What are GAIN, DocuNET and DK-DocRE? seems not introduced in the baseline section.}

\subsection{Baselines}
\label{sec:compared-baselines}
%We compare the following methods with our proposed CAST framework. 
%\bing{check the baselines in result tables, see which baselines are not mentioned here.}
\paragraph{Vanilla Baselines} This approach trains existing state-of-the-art RE models on  incompletely annotated data and serves as our baseline method. 
As stated earlier, we use \textbf{ATLOP} as the backbone model for the DocRE experiments. In addition to ATLOP, we compare \textbf{GAIN} \citep{zeng-etal-2020-double}, \textbf{DocuNET} \citep{zhang2021document}, and \textbf{KD-DocRE} \citep{tan2022document} as our vanilla baselines. 
These methods are top-performing methods on the Re-DocRED dataset. However, similar to ATLOP, the performances of these models deteriorate significantly under the incomplete annotation setting.

\paragraph{Negative Sampling (NS)}~\citep{li2020empirical} This method tackles the incomplete annotation problem through negative sampling. To alleviate the effects of false negatives, this method randomly selects partial negative samples for training. Such an approach can help to alleviate the detrimental effect of the false negative problem.

%\paragraph{Entropy-based Negative Sampling} \citep{li-etal-2022-rethinking} This method is a modified version of vanilla negative sampling, instead of discarding NA instances randomly, this method selects NA instances based on classification entropy.

\paragraph{Vanilla Self-Training (VST)} (\citealp{peng-etal-2019-distantly}; \citealp{jie-etal-2019-better}) VST is a variant of simple self-training.
In this approach, models are trained with $N$ folds, and all  pseudo-labels are directly combined with the original labels. Then, a new model is trained on the datasets with combined labels. 

%\paragraph{K-Fold Self-learning} \citep{jie-etal-2019-better} This baseline uses K-Fold training to tackle the false negative problem in named entity  

\paragraph{Class Re-balancing Self-Training (CREST)} \citep{wei2021crest} This algorithm is the most advanced baseline of class-imbalanced semi-supervised training, re-samples the pseudo-labels generated by models. However, this sampling strategy only considers the frequencies of the training samples, whereas our CAST considers the per-class performance on the development set.

\paragraph{SSR Positive Unlabeled Learning (SSR-PU)} \citep{wang2022unified} This method applies a positive unlabeled learning algorithm for DocRE under the incomplete annotation scenario. SSR-PU utilizes a shift-and-squared ranking (SSR) loss to accommodate the distribution shifts for the unlabeled examples. 

\paragraph{BioRE Baselines} For the BioRE experiments, we compare our methods with Biaffine Relation Attention Network \textbf{BRAN} \citep{verga-etal-2018-simultaneously} and \textbf{PubmedBERT} \citep{gu2021domain}, which is a pretrained language model in the biomedical domain.
% and \textbf{BRAN-PubmedBERT},  the combination of the aforementioned two methods. 

\begin{figure*}[ht]
\centering
\begin{subfigure}{.31\textwidth}
  \centering
  \includegraphics[width=\linewidth]{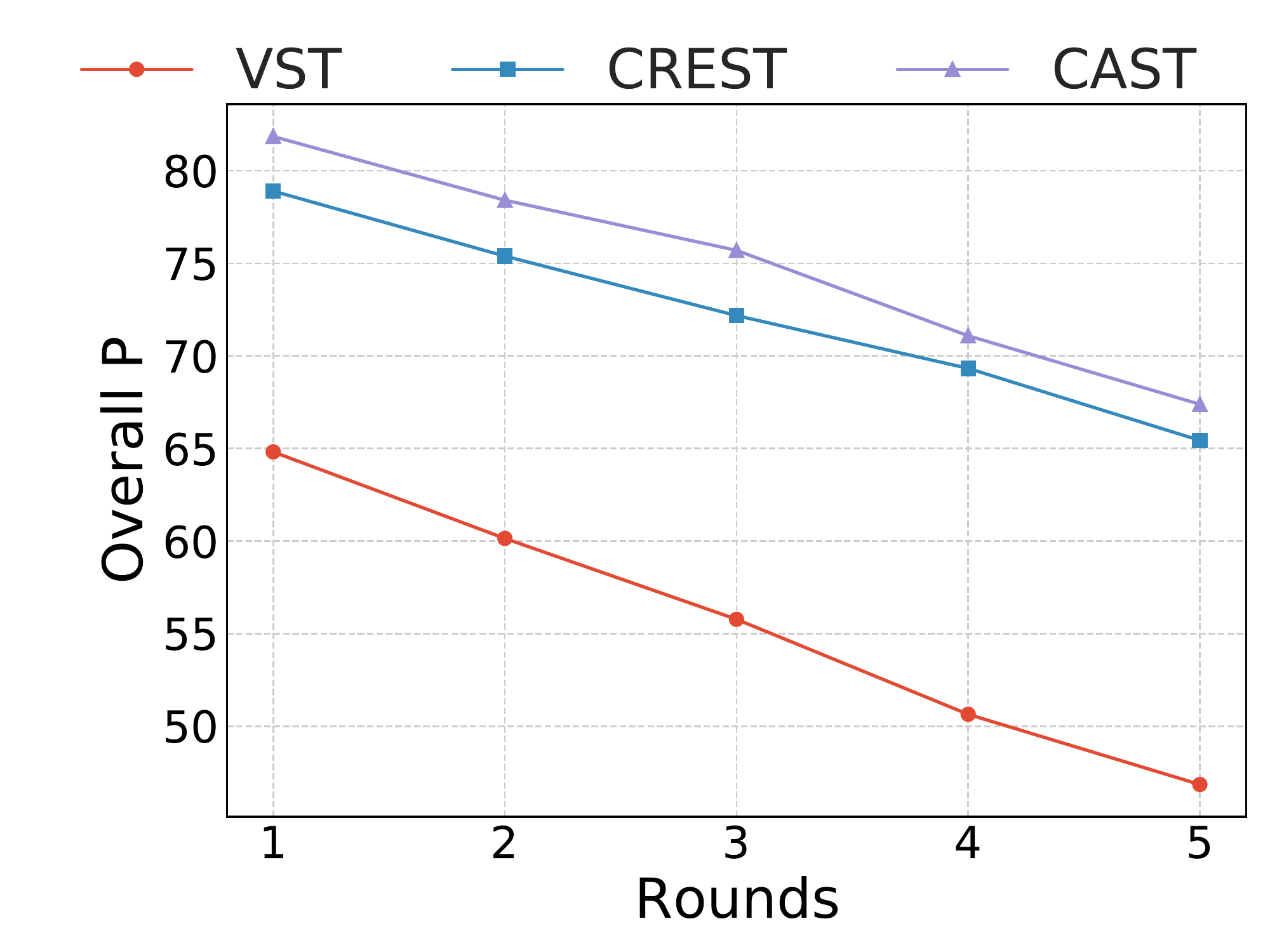}
  \caption{Precision}
  \label{fig:overall-p}
\end{subfigure}%
\begin{subfigure}{.31\textwidth}
  \centering
  \includegraphics[width=\linewidth]{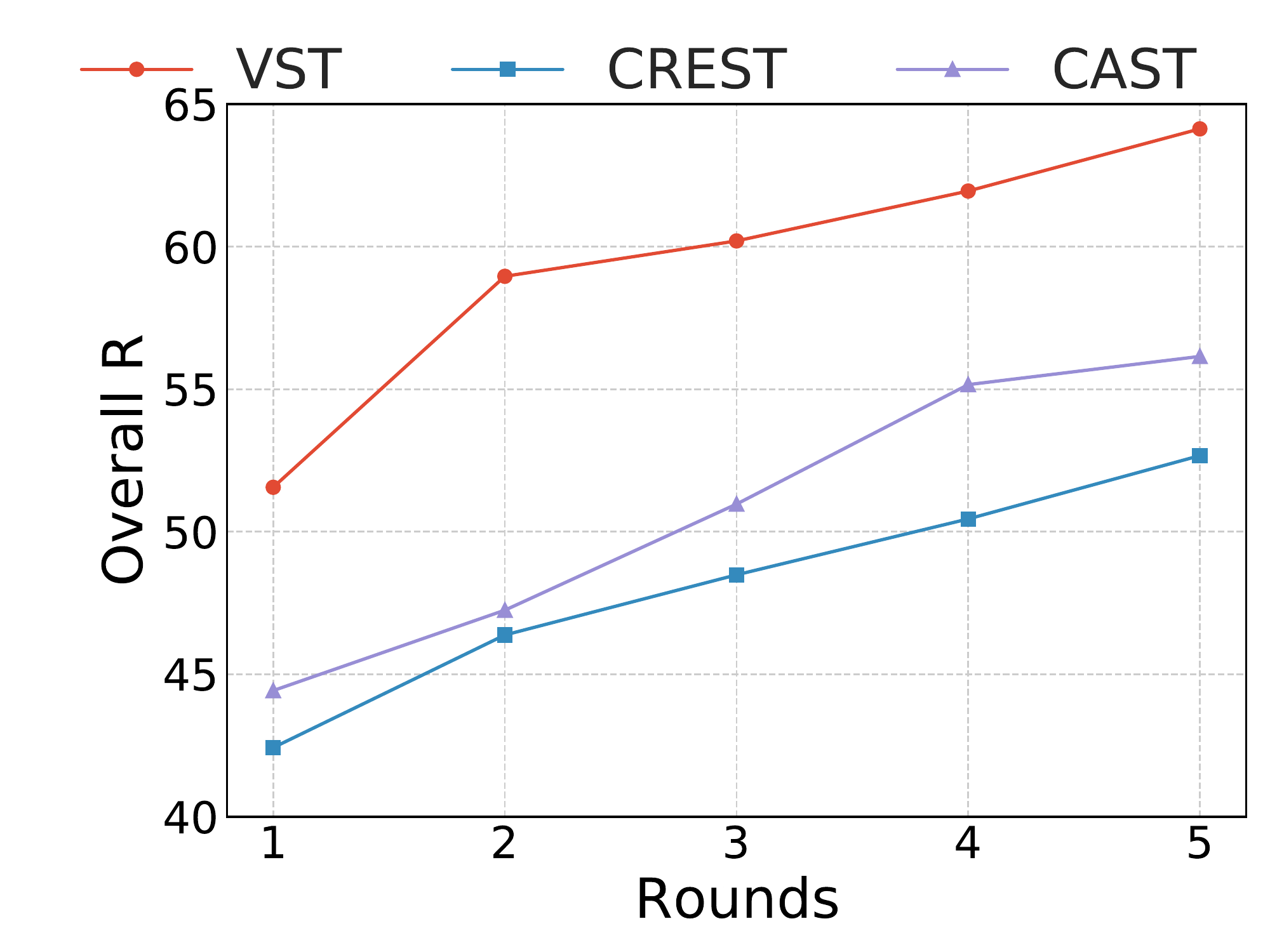}
  \caption{Recall}
  \label{fig:overall-r}
\end{subfigure}  % 
\begin{subfigure}{.31\textwidth}
  \centering
  \includegraphics[width=\linewidth]{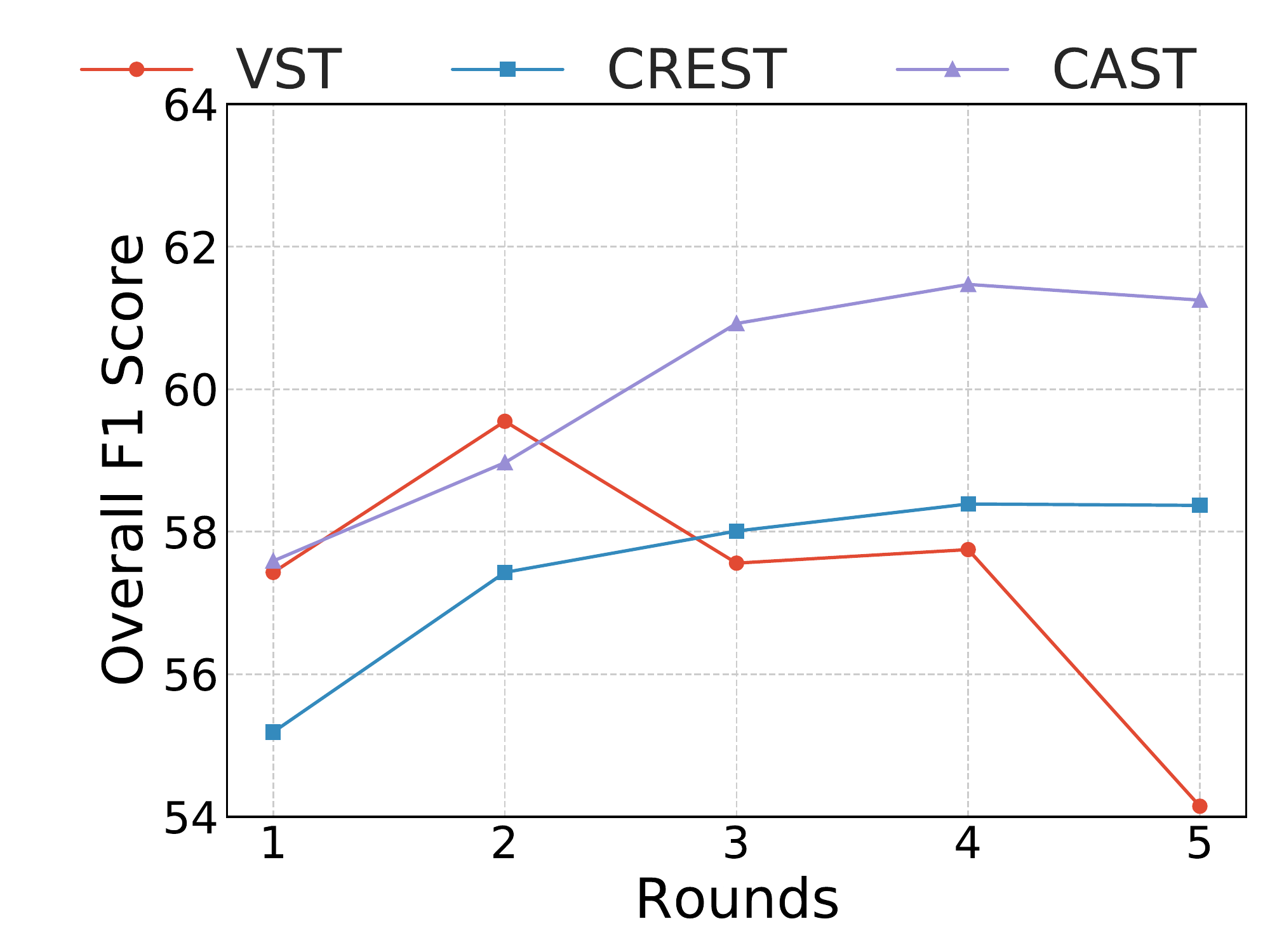}
  \caption{Overall F1}
  \label{fig:overall-f1}
\end{subfigure}%

\caption{Comparison of different self-training strategies when training on DocRED. }
\label{fig:self-training}
\end{figure*}
%\bing{there is an obvious error, at round 1, precision and recall of VST are both lower than CAST, how come its F1 is higher.}
%\bing{use bigger font size in the figures, but keep using small figures.}

\begin{table}[t]
\centering
\resizebox{\columnwidth}{!}{
\begin{tabular}{llrrrrrr} 
\toprule
 &\textbf{Model}                       & \multicolumn{1}{l}{\textbf{P}} & \multicolumn{1}{l}{\textbf{R}} & \multicolumn{1}{l}{\textbf{F1}} & \multicolumn{1}{@{~}l}{\textbf{Ign\_F1}} & \multicolumn{1}{@{}l}{\textbf{Freq\_F1}} & \multicolumn{1}{@{}r}{\textbf{LT\_F1}}  \\ 

\midrule
\multicolumn{4}{l}{\textit{Re-DocRED Training Data}}                      &                  &                  &                               \\
\midrule
\multirow{5}{*}{\rotatebox[origin=c]{90}{\textbf{BERT}} }
&ATLOP           & \textbf{86.70}	                & 62.46	               & 72.61	                  & 71.86                       & 75.92                        & 67.46                      \\
&NS-ATLOP       &    77.63	              &    69.17	           &      73.16	            &     72.92                  &       77.28                  &    67.59                  \\
&VST-ATLOP         & 72.77	                & \textbf{75.55}	                & 74.14	                & 72.48                          & 78.47                      & 68.13                       \\
& CREST-ATLOP                  & 75.94	              & 72.47	                 & 	74.17		                & 72.77	                         & 77.93                     & 68.68                       \\
\cmidrule(lr){2-8}
&\textbf{CAST-ATLOP} (Ours) & 76.59                 & 72.84                 & \textbf{74.67}                  & \textbf{73.32}                       &    \textbf{78.53}                     &    \textbf{69.34}                    \\

\bottomrule

\end{tabular}}
\caption{Experimental results on the test set of Re-DocRED when trained on silver quality data. }
\label{tab:silver-redocred}
\end{table}

\begin{table}[t]
\centering
\resizebox{\columnwidth}{!}{
\begin{threeparttable}

\begin{tabular}{llccc} 
\toprule
 &\textbf{Model}                & \multicolumn{1}{c}{\textbf{P}} & \multicolumn{1}{c}{\textbf{R}} & \multicolumn{1}{c}{\textbf{F1}}  \\ 
\midrule
&BRAN$^{\dag}$              & 41.8                  & 26.6                  & 32.5                    \\
\midrule
\multirow{7}{*}{\rotatebox[origin=c]{90}{\textbf{PubMedBERT}} }
&PubMedBERT$^{\dag}$        & 64.3                  & 31.3                  & 42.1                   \\
&BRAN$^{\dag}$   & 70.9                  & 31.6                  & 43.8                    \\
&ATLOP$^*$  & \textbf{76.17} \small{$\pm$ 0.36}                 & 29.70 \small{$\pm$0.54 }                  & 42.73 \small{$\pm$0.36}                   \\
&SSR-PU-ATLOP$^*$ & 54.27 \small{$\pm$0.23}                 & 43.93 \small{$\pm$0.40}                 & 48.56 \small{$\pm$0.32}                  \\ 
&NS-ATLOP     & 71.54 \small{$\pm$0.50}	 & 35.52 \small{$\pm$0.29}	  & 47.47 \small{$\pm$0.37}   \\
&VST-ATLOP   & 54.92 \small{$\pm$0.42}               & \textbf{48.39} \small{$\pm$0.58}               & 51.24 \small{$\pm$0.30}                   \\
&CREST-ATLOP  & 59.42 \small{$\pm$1.63}	  & 42.12 \small{$\pm$0.65}	  & 49.28 \small{$\pm$0.21}    \\
\cmidrule(lr){2-5}
&\textbf{CAST-ATLOP} (Ours)   & 66.68 \small{$\pm$2.22}                & 45.48 \small{$\pm$1.27}                 & \textbf{54.03} \small{$\pm$0.17}                  \\
\bottomrule
\end{tabular}
  % \begin{tablenotes}[para]
  %   \item[1] \citet{zhang-etal-2022-distant}  \item[2] \citet{wang2022unified}
  %   % \vspace{-0.3cm}
  % \end{tablenotes}
  \end{threeparttable}
}
\caption{Experimental results on  ChemDisGene. The results with numeric superscripts are taken from the respective papers. $^{\dag}$: The results are retrieved from \citet{zhang-etal-2022-distant}.
    $^*$: The results are retrieved from \citet{wang2022unified}.
    }
\label{tab:bio-re}
\end{table}

\begin{figure*}[t]
\begin{subfigure}{\textwidth}
  \centering
    \centering
    \resizebox{\textwidth}{!}{
    \includegraphics{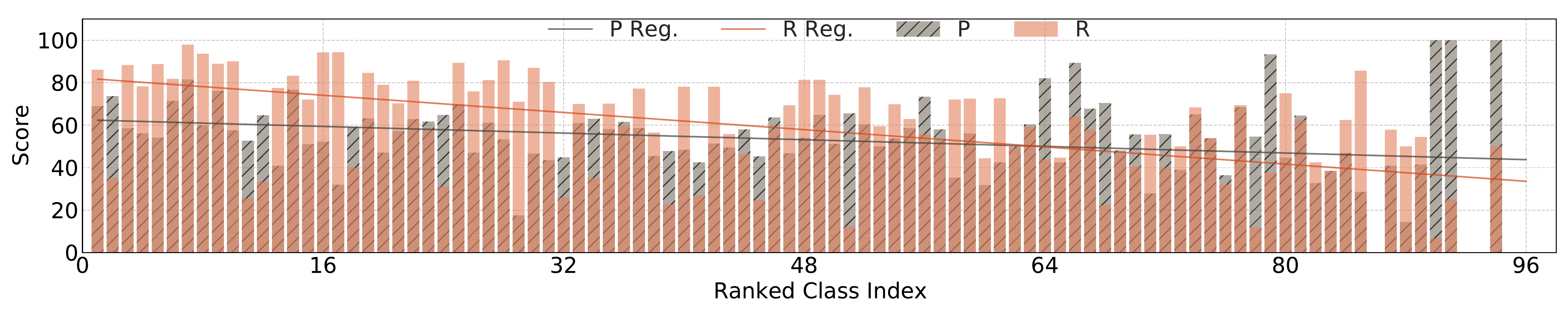}
    }
    \caption{VST}
    \label{fig:p-r-class-vst}
    
\end{subfigure}
\begin{subfigure}{\textwidth}
  \centering
    \centering
    \resizebox{\textwidth}{!}{
    \includegraphics{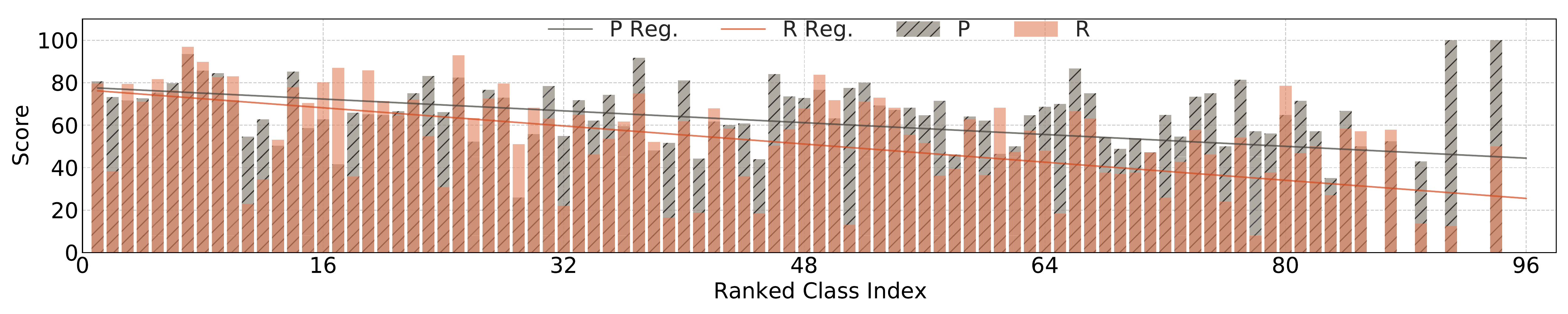}}
    \caption{CAST}
    \label{fig:p-r-class-cast}
\end{subfigure}
     \caption{Precision and recall scores for each class (ranked by class frequency: left (high)\textrightarrow right (low)) on the dev set of Re-DocRED of VST and CAST's best models, which are trained on DocRED. Better view in color.}

    \label{fig:p-r-class-compare}
\end{figure*}

\subsection{Experimental Results}
\label{sec:experiment-result}
%\bing{``, as shown in Figure \ref{fig:p-r-per-class}''?}
Table~\ref{tab:bronze-redocred} presents the experimental results for the document-level RE.
The experimental results on the original DocRED dataset show that the F1 score of the ATLOP-RoBERTa model is only 49.32. This finding can be ascribed to the low recall score of this method, as shown in Figure \ref{fig:p-r-per-class}. 
 NS significantly improves the performance compared with the baseline. 
 After comparing vanilla self-training with the baseline, we observe that although the recall score is the highest for this method, its precision is significantly reduced. 
 We observe similar trends for all self-training based methods (i.e., VST, CREST, and CAST), the recall improved at the expense of precision. 
 Notably, the performance of the simple NS baseline exceeds the performance of SSR-PU when trained on the DocRED data. 
Our proposed CAST framework consistently outperforms the competitive baselines and achieves the highest performance for both BERT and RoBERTa encoders. 
 Our best-performing model outperforms the baseline by 16.0 F1 (49.32 vs. 65.32). Moreover, the CAST obtains the highest precision score among the three self-training methods, thereby showing that the examples added by our class-adaptive sampling strategy have better quality.
 
The experimental results on the test set of Re-DocRED (Table~\ref{tab:silver-redocred}) depict that the baseline F1 score is significantly improved due to the large gain in the recall score when the training data are switched from bronze-quality to silver-quality. Compared with baseline approaches, our CAST achieves consistent performance improvements in terms of F1 score.
The F1 difference between the baseline and our CAST is 2.06 (72.61 vs. 74.67).
However, the performance gap between our approach and the baseline is smaller than the corresponding gap when both are trained with DocRED. 
This indicates that the performance of existing state-of-the-art models for document-level RE is decent when high-quality training data is provided but declines when the training data are incompletely annotated. This finding verifies the necessity of developing better self-training techniques because preparing high-quality training data is costly.
 
Table~\ref{tab:bio-re} presents  the experiments on biomedical RE. Our CAST model consistently outperforms strong baselines, exceeding the performance of SSR-PU by 5.47 F1 (54.03 vs. 48.56).
%\bing{need some discussion.}

On the basis of the results of DocRE and BioRE experiments, self-training-based methods aim to improve recall and consistently improve overall performance when the training data is incompletely annotated. However, our CAST maintains a better balance between increasing recall and maintaining precision.

\begin{figure}[t]
\centering
\begin{subfigure}{.5\columnwidth}
  \centering
  \includegraphics[width=\linewidth]{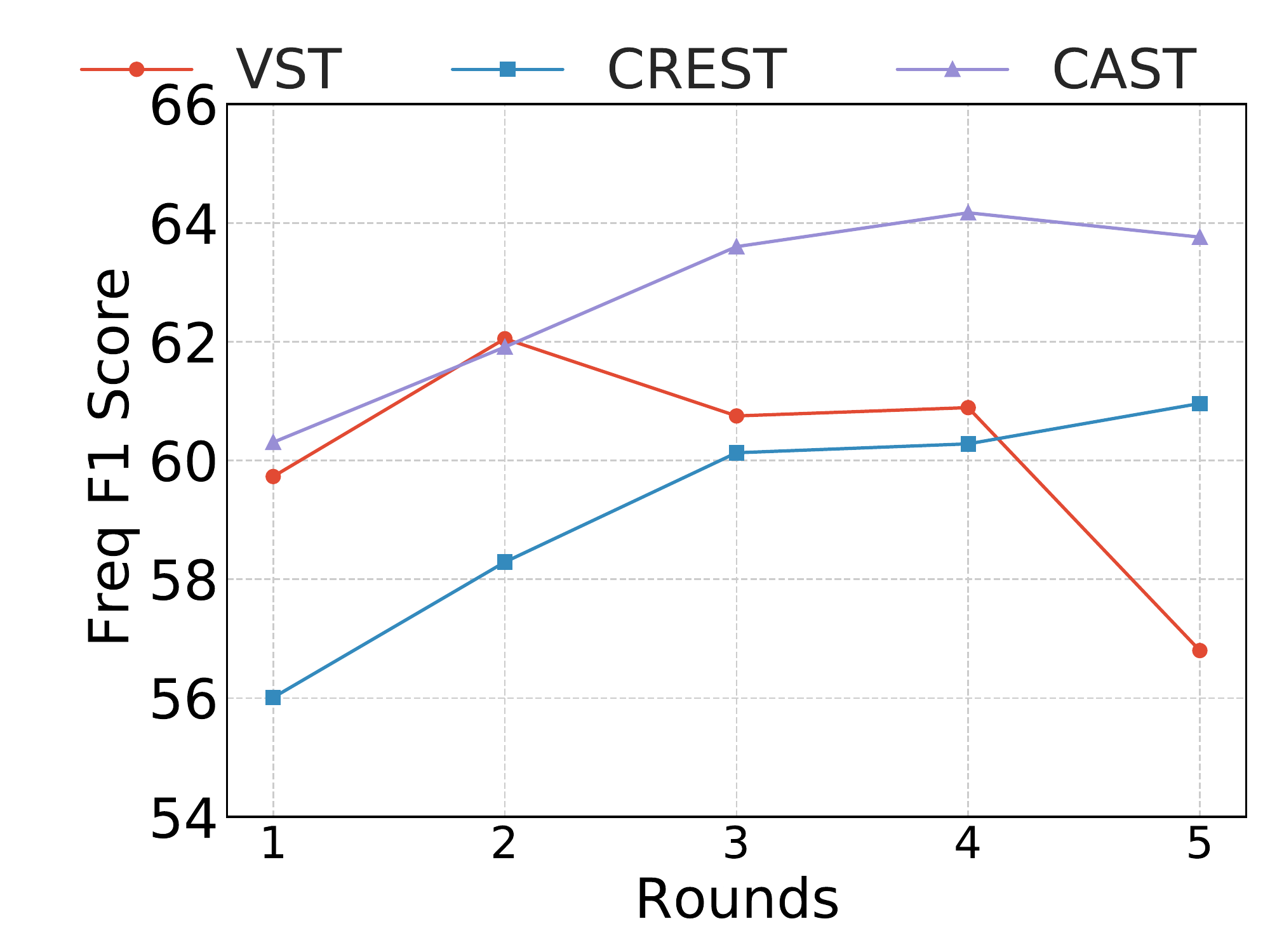}
  \caption{Frequent F1}
  \label{fig:freq-f1}
\end{subfigure}%
\begin{subfigure}{.5\columnwidth}
  \centering
  \includegraphics[width=\linewidth]{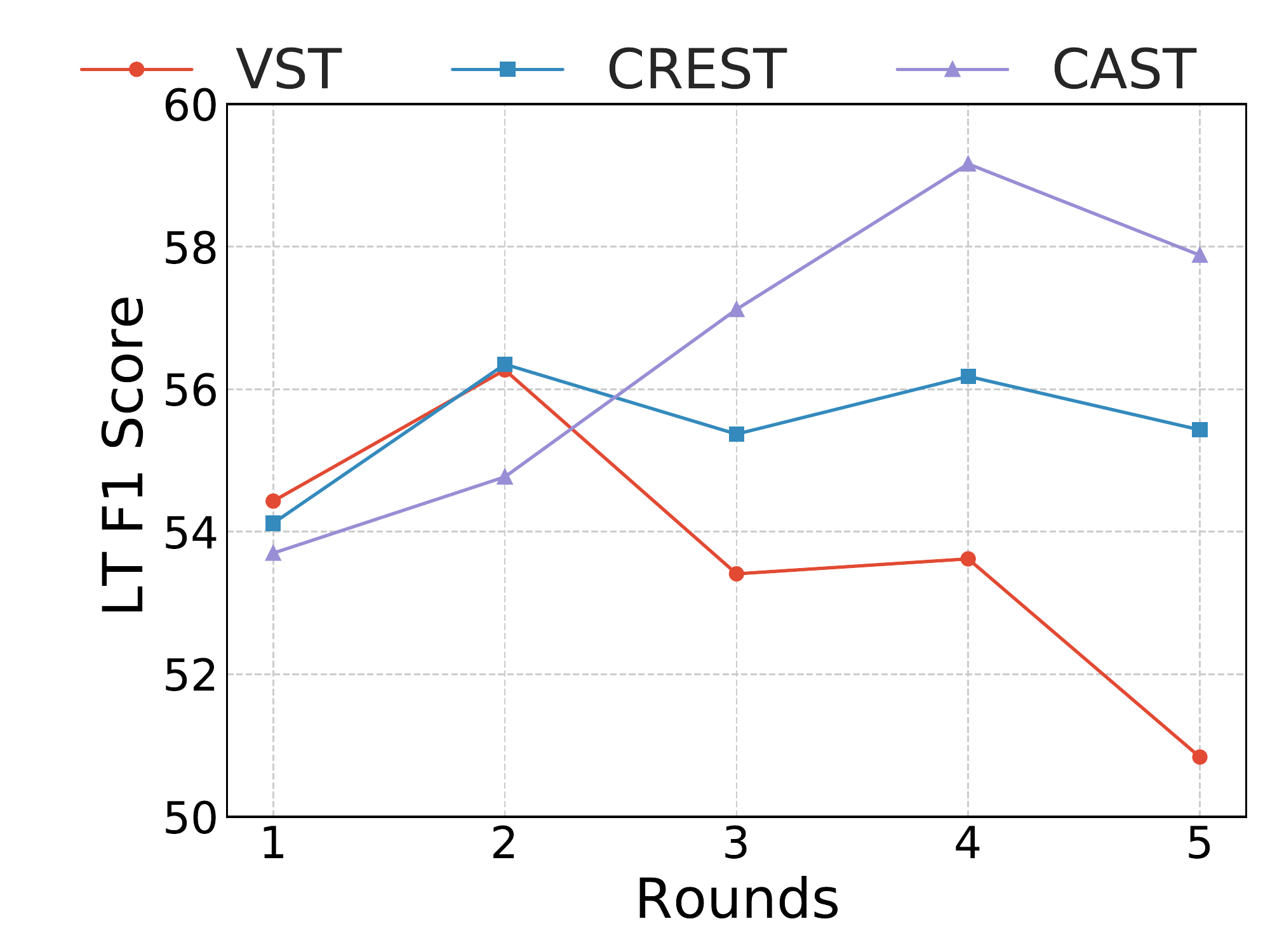}
  \caption{Long-tail F1}
  \label{fig:lt-f1}
\end{subfigure}  % 
\caption{F1 scores of frequent and long-tail classes with respect to rounds when trained on DocRED. }

\label{fig:freq-lt-f1}
\end{figure}

%\bing{use bigger font size in the figures, but keep using small figures.}

\begin{figure}[t]
    \centering
    \resizebox{0.9\columnwidth}{!}{
    \includegraphics{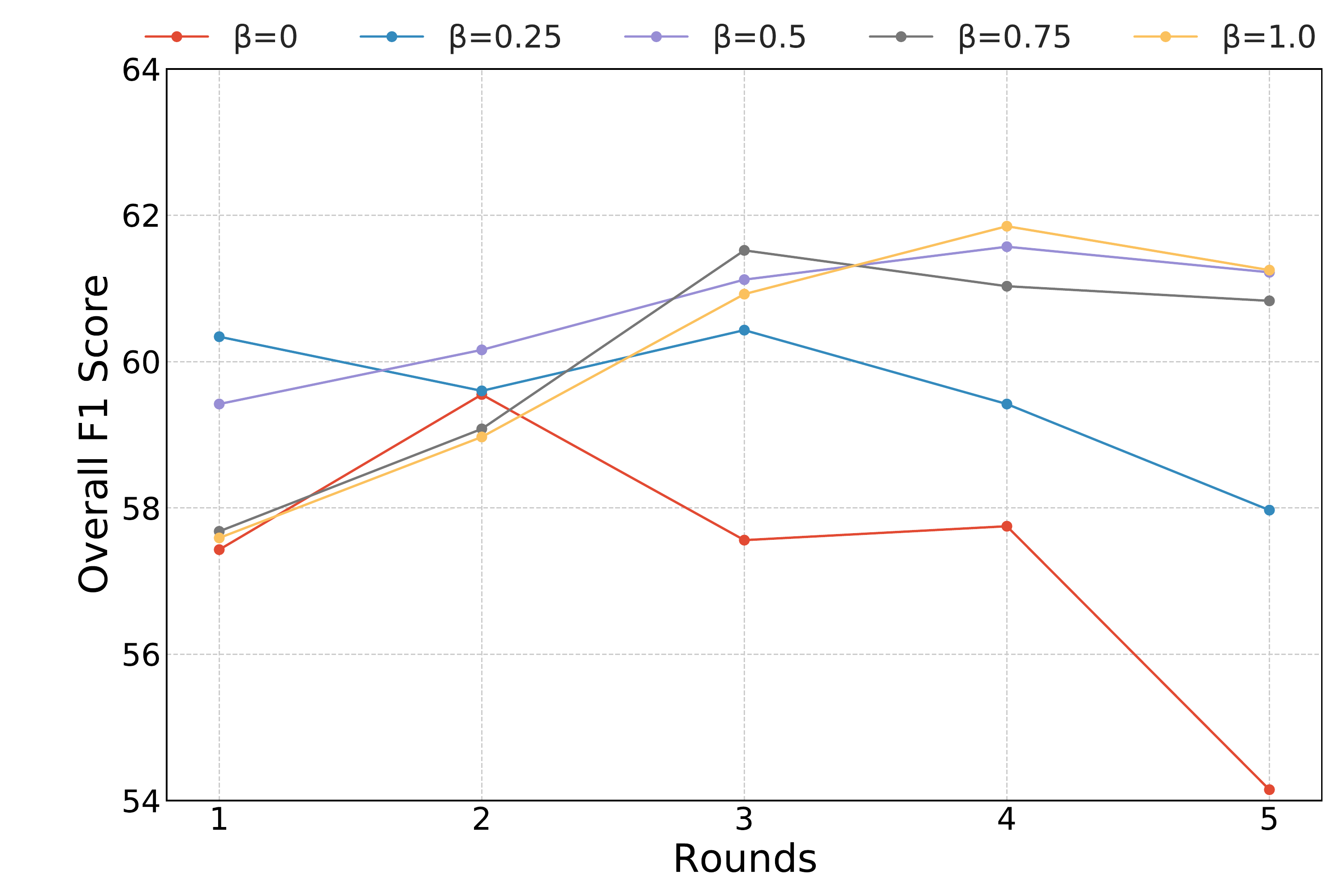}}
    \caption{Effect of different $\beta$ values. }
    \label{fig:beta-f1}
\end{figure}

%\bing{use bigger font size in the figures, but small figures.}

\section{Analysis}
\subsection{Comparisons of Self-Training Strategies}
To further compare different self-training strategies, we illustrate the detailed performance with respect to the self-training rounds in Figure~\ref{fig:self-training}. 
The reported scores are on the development set of Re-DocRED and the training data is from DocRED. 
Figure~\ref{fig:overall-r} shows that all self-training-based methods generally have improving  recall scores as the number of self-training rounds increases. On the contrary, the precision scores decline. 
From Figure~\ref{fig:overall-f1}, we observe that VST outperforms CREST and CAST in the first two rounds. This is mainly because VST does not perform re-sampling on the pseudo-labels and it utilizes all pseudo-labels. At the beginning stage, these labels are of relatively good quality. 
However, the performance of VST drops after the second round of pseudo-labeling because as the number of rounds increases, the increase in the number of false positive examples in the pseudo-labels outweighs the benefit.
Meanwhile, the performance gains of CREST and CAST are relatively stable, and both methods produce their best-performing models at round 4. Compared with CREST, our CAST maintains higher precision scores as the number of rounds increases (Figure~\ref{fig:overall-p}).

We also assess the F1 performance of the frequent and long-tail classes with respect to the number of rounds, and the comparison is shown in Figure~\ref{fig:freq-lt-f1}.
The results reveal that VST suffers greatly from confirmation bias on both frequent and LT classes, i.e., Figure~\ref{fig:freq-f1} and Figure~\ref{fig:lt-f1}, and its performance becomes very poor in round 5. In Figure \ref{fig:lt-f1}, we can see that the performance gains of CAST is stable across the training rounds and achieved the best LT performance.

\subsection{Detailed Analysis of CAST}
In this section, we analyze the performance of our CAST framework in detail. 
We first plot the precision and recall scores of VST and CAST for all the classes in Figure~\ref{fig:p-r-class-compare}, where the experimental results are obtained by training with the DocRED dataset. The formulation of Figure~\ref{fig:p-r-class-compare} is the same as Figure~\ref{fig:p-r-per-class}. Figure~\ref{fig:p-r-class-vst} demonstrates that VST significantly improves the recall scores of many classes compared with the baseline in Figure~\ref{fig:p-r-per-class}. However, the improvements in recall scores are accompanied by a large decline in precision scores. This observation shows that the pseudo-labels in VST contain a considerable amount of erroneous predictions. By contrast, our CAST framework is able to better maintain the precision scores for most of the classes. The recall scores for most of the classes are significantly higher compared with those of the baseline. This observation justifies the improvements of the overall F1 scores in Table~\ref{tab:bronze-redocred} despite the lower recall of CAST model than VST.

%\bing{need to be specific, which training data, which task.}
\subsection{Effect of $\beta$}
We further analyze the effect of the sampling coefficient $\beta$ on our CAST framework in Figure~\ref{fig:beta-f1}, the experiments are conducted by training with the DocRED dataset. 
When $\beta$ value is small, CAST behaves like the VST model, exhibits some F1 improvements in the first few rounds, and demonstrates diminishing positive effects in the later rounds. 
Larger $\beta$ leads to better overall improvements and smaller fluctuations across different rounds. However, because the term $[P_{i} * (1 - R_{i})]$ in Eq.~\ref{eqn:p-r-prob} is smaller than 1, higher $\beta$ may lead to lower sampling rates for all the classes.
As a result, the convergence time of self-training may be longer. 
The interpretation for other values of $\beta$ is provided in the Appendix~\ref{sec:large-beta}.

%Note that $\beta > 1.0$ can potentially achieve better results because it carries out a more cautious but accurate sampling in each step. However, it also increases the computation cost.

%\item frequent class vs rare classes performance figs
%\item performance w.r.t number of rounds M, N (merge with prior)
%\item precision recall hyper-parameter ()
\section{Conclusions and Future Work}
In this work, we study the under-explored problem of learning from incomplete annotation in relation extraction.
This problem is highly important in real-world applications. We show that existing state-of-the-art models suffer in this scenario. 
To tackle this problem, we proposed a novel CAST framework. We conducted experiments on DocRE and BioRE tasks, and experimental results show that our method consistently outperforms competitive baselines on both tasks. For future work, we plan to extend our framework to the distant supervision scenario. From the domain perspective, we plan to apply our framework to image classification tasks.

\section{Limitations}
The proposed CAST framework carries the same limitation of self-training-based methods, which is the requirement for multiple rounds and multiple splits of training. As a result, the GPU computing hours of CAST are longer than those of vanilla baselines and NS. 
% However, the self-training based methods are all susceptible to this limitation.

\bibliography{custom}
\bibliographystyle{acl_natbib}

\appendix
\begin{figure*}[ht]
\centering
\begin{subfigure}{.3\textwidth}
  \centering
  \includegraphics[width=\linewidth]{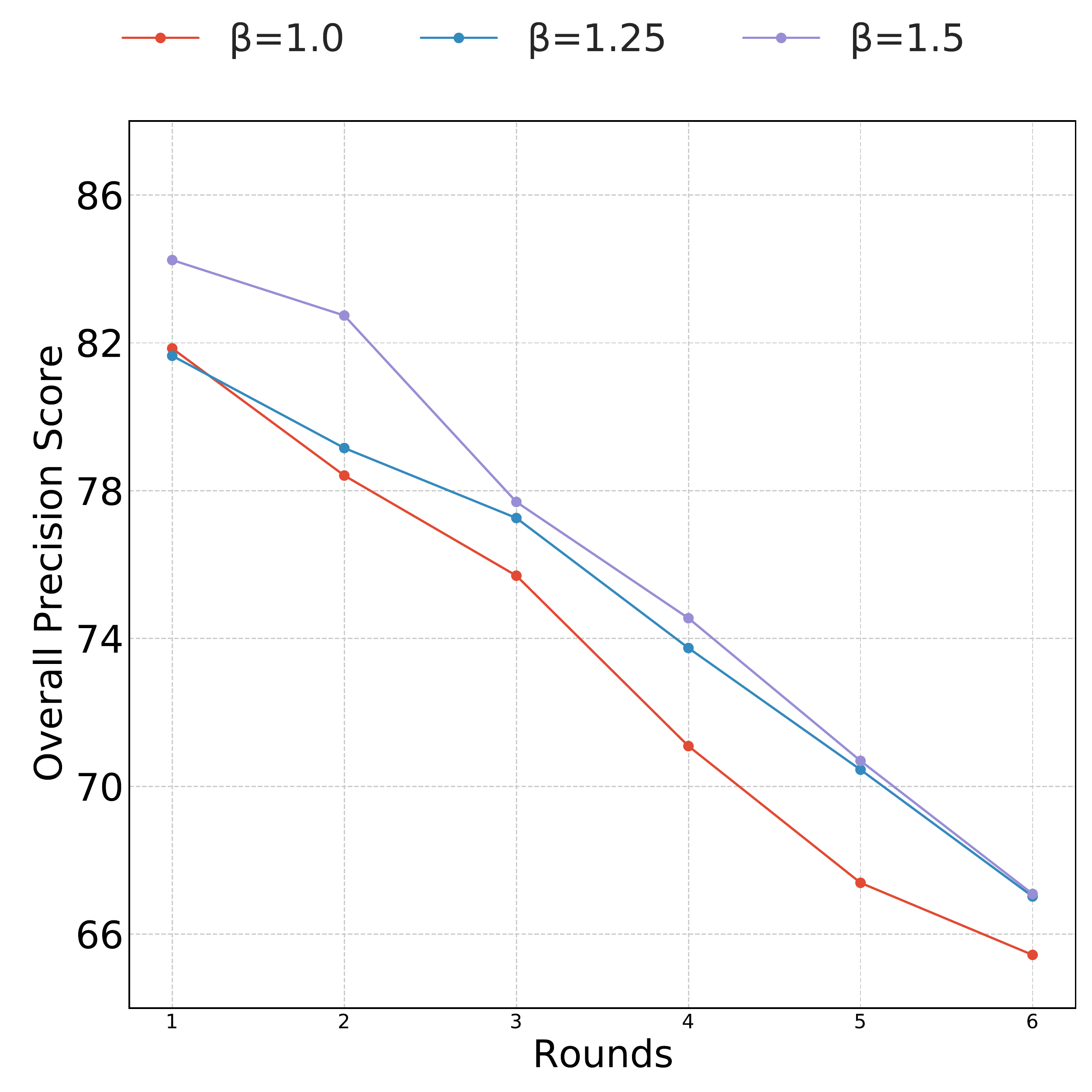}
  \caption{Precision}
  \label{fig:overall-p-large-b}
\end{subfigure}%
\begin{subfigure}{.3\textwidth}
  \centering
  \includegraphics[width=\linewidth]{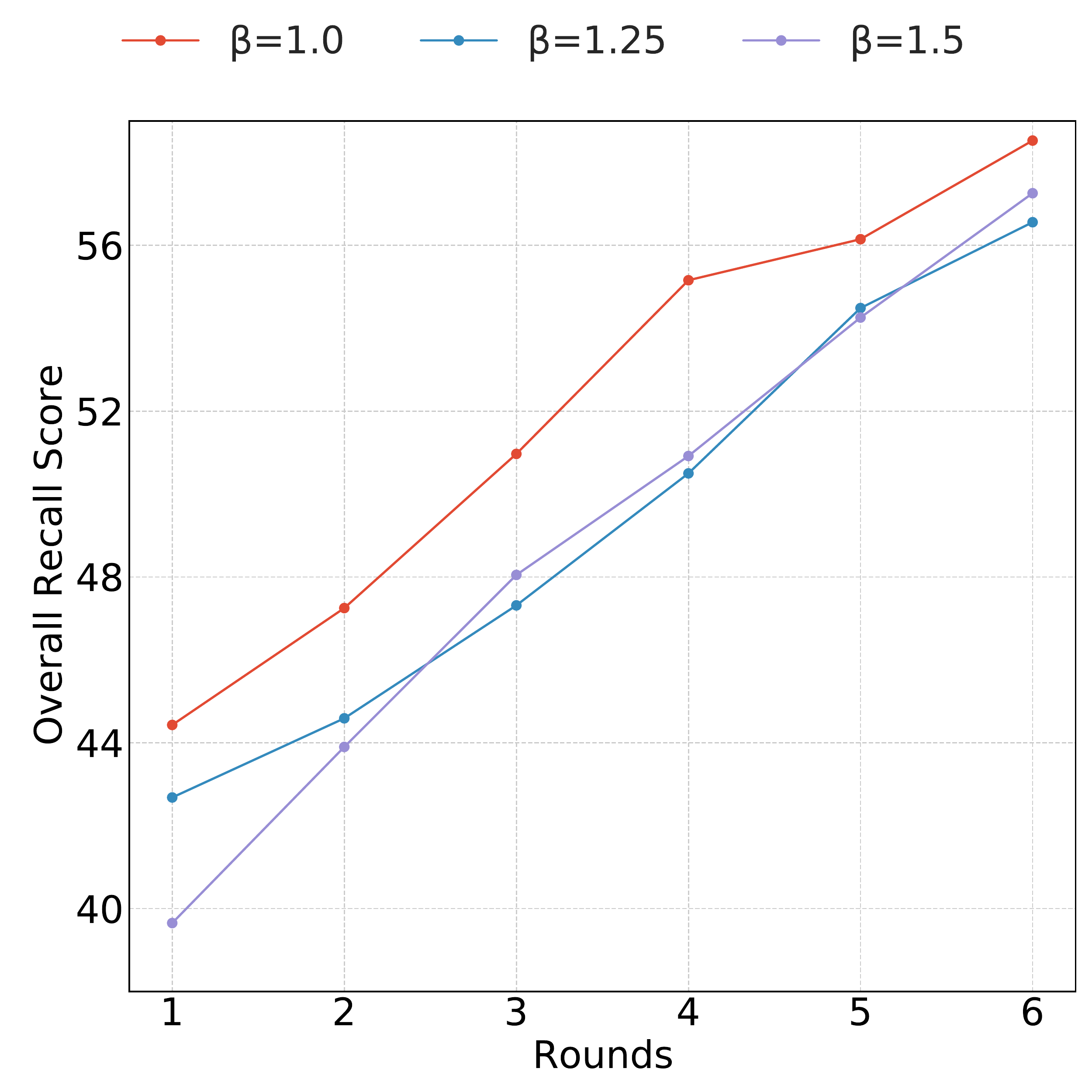}
  \caption{Recall}
  \label{fig:overall-r-large-b}
\end{subfigure}  % 
\begin{subfigure}{.3\textwidth}
  \centering
  \includegraphics[width=\linewidth]{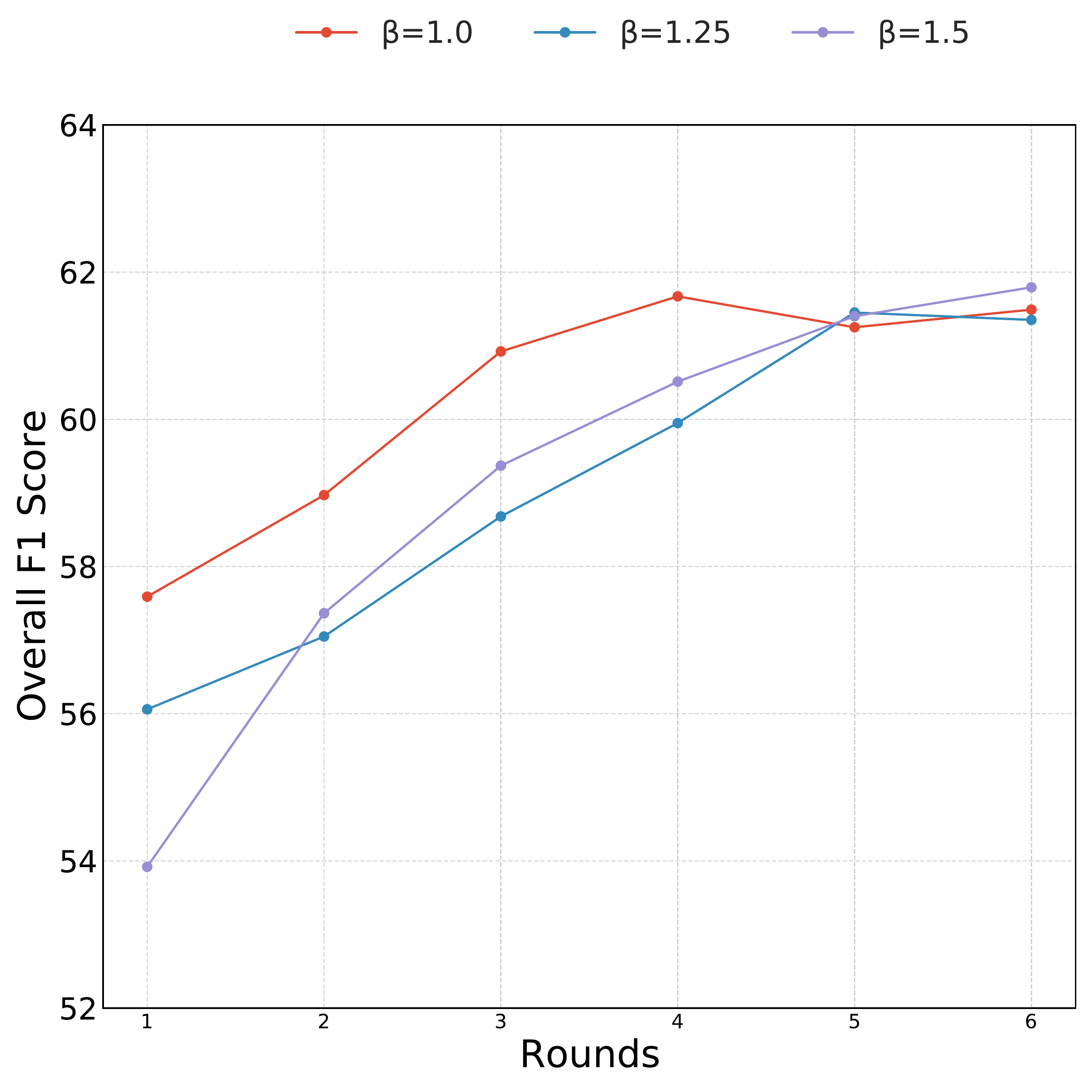}
  \caption{Overall F1}
  \label{fig:overall-f1-large-b}
\end{subfigure}%

\caption{Effect of larger $\beta$ when training on DocRED. }
\label{fig:self-training-large-b}
\end{figure*}
\section{Sentence-Level Relation Extraction}
%\bing{add reference to the datasets.}
Besides document-level RE, we also examined our method for sentence-level relation extraction (SentRE), the task is a simplified version of its document-level counterpart. Compared to the DocRE setting, there are two main differences for sentence-level RE. 
First, there are exactly $n=2$ entities for each SentRE example. 
Second, there is only one relation type for an entity pair in SentRE, whereas there can be multiple relation types for DocRE. 
Again, we used two types of training data for the SentRE task. 
The first set of training data is from the original TACRED dataset, and the second set of training data is from Re-TACRED. 
Compared to the revision of Re-DocRED, which only resolved the false negative problem\footnote{The problem of false positive is minor in DocRED.}, the revision of Re-TACRED not only resolved the false negative problem but also relabeled the false positive instances.

The experimental results on SentRE are shown in Table~\ref{tab:re-tacred}. For the TACRED dataset, the top 5 classes\footnote{They cover 57.5\% of the positive instances.} are included in the frequent classes.  We can also see that when training with bronze-quality data (i.e., the upper section), our proposed CAST still achieves the best performance in terms of F1 score. 
This observation shows that our method is effective across different relation extraction scenarios and backbone models. 
On the other hand, we can observe that the baseline model achieves the highest F1 score when training with the Re-TACRED dataset (i.e., the lower section). 
As mentioned in the section of problem definition, the Re-TACRED training set has resolved the false negative and false positive problems of TACRED. 
Therefore, by simply using all training samples of Re-TACRED, the baseline approach achieves the best F1. It is worth noting that our CAST is very robust and does not hurt the performance, i.e., achieving slightly worse F1 but slightly better recall compared with the baseline.

\begin{table}[ht]
\centering
\resizebox{0.9\columnwidth}{!}{
\begin{tabular}{lccccc} 
\toprule

 Model                    & \multicolumn{1}{c}{\textbf{P}} & \multicolumn{1}{c}{\textbf{R}} & \multicolumn{1}{c}{\textbf{F1}} & \multicolumn{1}{@{}c}{\textbf{Freq\_F1}} & \multicolumn{1}{@{}c}{\textbf{LT\_F1}}  \\ 
\midrule
\multicolumn{4}{l}{\textit{TACRED Training Data}}                                                             &                      &                        \\
\midrule
\textbf{Baseline}             & \textbf{80.64}                 & 40.44                 & 53.87                  & 70.62                        & 36.43                       \\
\textbf{NS}    & 62.37                 & \textbf{53.96}                 & 57.86                  & 74.49                        & 40.57                       \\
\textbf{VST}        & 67.24                 & 52.83                 & 59.17                  & 80.64                        & 47.25                       \\
\textbf{CREST}                & 67.92                 & 52.48                 & 59.21                  & 80.36                        & 47.64                       \\
\textbf{CAST (Ours)}       & 73.33                 & 51.03                 & \textbf{60.18}                  & \textbf{81.12}                        & \textbf{48.75}                       \\
%\textbf{HNM}                  & 64.86                 & 52.99                 & 58.33                  & 78.93     & 47.13                       \\
%\textbf{CAST + HNM} & 71.58                 & 52.75                 & \textbf{60.74}                  & \textbf{81.63}                        & \textbf{49.24}                       \\
\midrule
\multicolumn{4}{l}{\textit{Re-TACRED Training Data}}                                                              &                      &                        \\
\midrule
\textbf{Baseline}             & \textbf{88.01}                & 	87.82               & 	\textbf{87.91}	              & \textbf{89.21}        & 	\textbf{87.37}\\
\textbf{NS}              & 	 85.44                & 	 88.56                 & 	 86.97                & 88.75 	   & 86.24  \\ 
\textbf{VST}             & 83.71	                & \textbf{89.82}	                 & 86.65	                & 87.94	   &  85.99 \\ 
\textbf{CREST}               & 86.46 	                & 88.45 	                 & 87.45 	                & 88.64 	   &  86.97\\ 

\textbf{CAST (Ours)}             & 87.62	                & 	87.96	                 & 	87.78		                & 89.14	  &  87.31 \\ 
\bottomrule
\end{tabular}}
\caption{Experimental results on the test set of Re-TACRED when trained on TACRED and Re-TACRED, respectively. Model selection is based on the dev set of Re-TACRED. }
\label{tab:re-tacred}
\end{table}

\section{Hyper-Parameters of the Baselines}
In this section, we report the hyper-parameters of the baseline experiments. For the negative sampling experiments, we used sampling rate $\gamma = 0.1$ for the DocRED experiment, $\gamma = 0.5$ for TACRED experiment and $\gamma = 0.7$ for the Re-TACRED and Re-DocRED experiments. $\gamma$ is searched from $\gamma \in \{0.1, 0.3, 0.5, 0.7, 0.9\}$. 

From CREST~\citep{wei2021crest}, the classes are first ranked by their frequencies, and the sampling rate for class $i$ is calculated as:
\begin{equation}
    \mu_{i} = (\frac{X_{|C| + 1 - i}}{X_{1}})^{\alpha}
\end{equation}
where $X_{1}$ is the count of the most frequent class among the positive classes. We set the power $\alpha = 0.33$ as reported in their paper. For all the self-training-based experiments (VST, CREST, and CAST), we trained with 10 epochs per fold. All our experiments were run on a NVIDIA-V100 GPU.

\section{Experiments on larger $\beta$}
\label{sec:large-beta}
In Figure~\ref{fig:self-training-large-b}, we show the experimental results when $\beta$ is larger than 1.0. Increasing $\beta$ inevitably reduces the sampling probability for all the classes, which is more conservative. Therefore, larger $\beta$ tends to have higher precision scores and lower recall scores. From Figure~\ref{fig:self-training-large-b}, we see that the optimal round for F1 scores is 4 for $\beta = 1.0$ and 5 for $\beta = 1.25$. When $\beta > 1.5$, the F1 score may not reach the optimal point before the 6th round. Since CAST would require training $MN$ times, larger $\beta$ may lead to significantly longer computation time to reach the optimal F1 score.   

\begin{table}[t]
\centering
\resizebox{\columnwidth}{!}{
\begin{tabular}{llrrrrrr} 
\toprule
 &\textbf{Model}                       & \multicolumn{1}{l}{\textbf{P}} & \multicolumn{1}{l}{\textbf{R}} & \multicolumn{1}{l}{\textbf{F1}} & \multicolumn{1}{@{~}l}{\textbf{Ign\_F1}} & \multicolumn{1}{@{}l}{\textbf{Freq\_F1}} & \multicolumn{1}{@{}r}{\textbf{LT\_F1}}  \\ 

\midrule
\multicolumn{4}{l}{\textit{DocRED Training Data with Incomplete $S_{D}$}}                      &                  &                  &                               \\
\midrule
\multirow{6}{*}{\rotatebox[origin=c]{90}{\textbf{BERT}} }
&ATLOP           & \textbf{88.39}		                & 28.87	             & 43.52		                  &    43.28                     &    45.49	                  & 40.46                         \\
&SSR-PU           & 70.42			                & 46.67	             & 56.14			                  &    55.21	            &    59.38     	                  & 	49.24                         \\
&NS-ATLOP       &    55.98		              &   55.63		           &      55.78		            &     53.90	                 &       58.73	              &    51.92                \\
&VST-ATLOP         &63.03	                &	\textbf{51.60}             & 56.71	                & 		         55.26                     &	    60.75            & 		51.52                        \\
& CREST-ATLOP                  & 72.83	              & 47.81                 & 		57.72	                & 	56.71	                       &59.05		                      & 54.82                      \\
\cmidrule(lr){2-8}
&\textbf{CAST-ATLOP} (Ours) & 70.97                 & 	50.70	                & \textbf{59.14}              & 	\textbf{58.03}	                   &    \textbf{61.20}	                       &    \textbf{56.22}                   \\

\bottomrule

\end{tabular}}
\caption{Experimental results on the test set of Re-DocRED when trained on DocRED training data and using the development set of DocRED for model selection. }
\label{tab:noisy-dev}
\end{table}

\section{Experiments with Incomplete $S_{D}$}
\label{sec:doc-dev-exp}
In this section, we conducted experiments on DocRED with a development set of lower quality. Specifically, we used $S_{D}$ from the DocRED dataset instead of Re-DocRED. The experiment results are shown in Table~\ref{tab:noisy-dev}. We can see that the over performances of most methods were decreased. This observation showed the importance of a high-quality development set when training with incomplete data. Nevertheless, our CAST model still achieves the best overall performance among the compared methods.

\end{document}